\documentclass{article}
\usepackage{todonotes}
\usepackage[utf8]{inputenc}
\usepackage{times}\usepackage{url}\usepackage[hidelinks]{hyperref}
\usepackage[small]{caption}
\usepackage{graphicx}
\usepackage{amsmath,amssymb}
\usepackage{booktabs}
\usepackage{tikz}
\usepackage{pgfplots}
\usepackage{comment}
\usepackage{algpseudocode,algorithm}
\newcommand{\AF}{\mathcal{F}}
\newcommand{\A}{\mathcal{A}}
\newcommand{\Adj}{\mathbb{A}}
\newcommand{\HC}{\mathbb{HC}}
\newcommand{\D}{\mathcal{D}}
\newcommand{\Att}{\mathtt{Att}}

\algnewcommand\algorithmicswitch{\textbf{switch}}
\algnewcommand\algorithmiccase{\textbf{case}}
\algnewcommand\algorithmicassert{\texttt{assert}}
\algnewcommand\Assert[1]{\State \algorithmicassert(#1)}%
\algdef{SE}[SWITCH]{Switch}{EndSwitch}[1]{\algorithmicswitch\ #1\ \algorithmicdo}{\algorithmicend\ \algorithmicswitch}%
\algdef{SE}[CASE]{Case}{EndCase}[1]{\algorithmiccase\ #1:}{\algorithmicend\ \algorithmiccase}%
\algtext*{EndSwitch}%
\algtext*{EndCase}%

\newtheorem{definition}{Definition}
\newtheorem{proposition}{Proposition}
\newtheorem{example}{Example}
\newtheorem{property}{Property}
\newtheorem{conjecture}{Conjecture}

\newenvironment{proof}{\textbf{Proof}}{\hfill$\square$}

\title{The Inverse Problem for Argumentation Gradual Semantics}

\author{Nir Oren$^1$\footnote{Contact Author} \and Bruno Yun$^1$\and Srdjan Vesic$^2$ \and Murilo Baptista$^1$\\
$^1$University of Aberdeen\\
$^2$CRIL\\
\{n.oren,b.yun,murilo.baptista\}@abdn.ac.uk, vesic@cril.fr}
\date{}

\begin{document}

\maketitle

\begin{abstract}
Gradual semantics with abstract argumentation provide each argument with a score reflecting its acceptability, i.e. how ``much'' it is attacked by other arguments.
Many different gradual semantics have been proposed in the literature, each following different principles and producing different argument rankings. 
A sub-class of such semantics, the so-called \emph{weighted semantics}, takes, in addition to the graph structure, an initial set of weights over the arguments as input, with these weights affecting the resultant argument ranking.
In this work, we consider the inverse problem over such weighted semantics. That is, given an argumentation framework and a desired argument ranking, we ask whether there exist initial weights such that a particular semantics produces the given ranking. 
The contribution of this paper are: (1) an algorithm to answer this problem, (2) a characterisation of the properties that a gradual semantics must satisfy for the algorithm to operate, and (3) an empirical evaluation of the proposed algorithm.
\end{abstract}

\section{Introduction}

Abstract argumentation semantics aim to identify the justification status of arguments by considering interactions between arguments. Such semantics typically operate over a directed graph, with nodes representing the (abstract) arguments, and directed edges denoting the interactions between them, e.g., attacks or supports among others. 
Standard argumentation semantics \cite{baroni_introduction_2011,dung_acceptability_1995,caminada_semi-stable_2012}  identify sets of arguments which are considered justified (as well as unjustified and undecided).
In contrast, ranking-based semantics seek to assign a ranking (or ordering) over arguments, with higher ranked arguments being considered more justified (or ``less attacked'') than lower ranked arguments.
Such rankings are --- in most ranking-based semantics --- determined by assigning numerical values (called \emph{acceptability degrees}) to all arguments, with the ranking on arguments being computed based on the numerical ordering.
Those ranking-based semantics are called \textit{gradual semantics}. 
Note that not all ranking-based semantics follow this numerical approach. For instance, the ranking on arguments obtained from the burden-based or the discussion-based semantics, defined in \cite{amgoud_ranking-based_2013}, are computed using the lexicographical order on vectors of argument scores.

While some ranking-based semantics \cite{amgoud_ranking-based_2013,bonzon_comparative_2016,delobelle_ranking-based_2017,amgoud_ranking_2016} only consider the structure of a standard Dung's argumentation framework, others take in one or more additional features, such as a set of \emph{initial weights} for each argument \cite{TB-sem,AMGOUD2022103607}; weights for attacks between arguments \cite{coste-marquis_weighted_2012,yun_2021_gradual}; a support relation \cite{mossakowski_modular_2018,mossakowski_bipolar_2016,rago_discontinuity-free_2016}; or  even set attacks \cite{yun_ranking-based_2020}. In most gradual semantics, the final acceptability degree of an argument then depends on a range of parameters. In this paper, we focus on gradual semantics which take into account the structure of the graph, the initial weights of arguments, and the peculiarities of the semantics being used. Of course, the proposed approach could easily be generalised to other settings.

Rather than describing how an initial set of argument weights map to a ranking on arguments via some semantics, in this paper we consider the inverse problem. That is, \emph{given an abstract argumentation framework and a desired ranking on arguments, we seek to identify what initial weights should be assigned to arguments} so as to obtain the desired argument ranking. We provide an algorithm for undertaking this task for a set of well-known gradual semantics (trust-based, iterative-schema, weighted max-based, weighted card-based, and weighted h-categorizer semantics) which satisfy some basic properties, and then empirically evaluate the algorithm's performance.

While we do not discuss the applications of our results, we note that potential areas in which they can be used include persuasion \cite{POLBERG2018487} and preference elicitation \cite{mahesar18computing}.

The remainder of this paper is structured as follows. First, in Section \ref{sec:background}, we provide the necessary background to understand our approach. Second, in Section \ref{sec:inverseProb}, we describe our algorithm. In Section \ref{sec:prop}, we highlight the requisite properties of the semantics over which the algorithm operates. Our empirical evaluation is detailed in Section \ref{sec:eval}, and we discuss potential applications of this work as well as avenues for future research in Section \ref{sec:discussion}.

\section{Background} \label{sec:background}

We begin this section by providing a brief overview of abstract argumentation, as well as  several gradual semantics. Following this, we introduce the bisection method, a simple technique for finding the roots of an equation which lies at the heart of our approach.

\subsection{Argumentation}

We situate our approach in the context of \emph{abstract argumentation}. Here, arguments are atomic entities which interact with each other via a binary attack relationship. Such systems are easily encoded as directed graphs (c.f., \cite{dung_acceptability_1995}). In this paper, since our departure point involves assigning each argument an initial weight, we instead consider \emph{weighted argumentation frameworks} (WAFs) \cite{dunne_weighted_2011,amgoud_acceptability_2017}.

\begin{definition}[WAF]
A weighted argumentation framework (WAF) is a triple $\AF=\langle \A,\D,w \rangle$, where $\A$ is a finite set of arguments, $\D \subseteq \A \times \A$ is a binary attack relation, and $w: \A \to [0,1]$ is a weighting function assigning an initial weight to each argument.
\end{definition}

The set of attackers of an argument $a \in \A$ is denoted as $\Att(a)=\{b \in \A \mid (b,a) \in \D\}$.

A ranking-based semantics allows us to move from a weighted argumentation framework  to a ranking over arguments. While myriad semantics have been proposed, we consider the gradual semantics described in \cite{AMGOUD2022103607} due to this work's recency and the popularity of the semantics described therein. We note in advance that some of these semantics do not work with our approach, but we will use these to help explain the properties of those semantics to which our approach applies. 
Furthermore, while \cite{bonzon_comparative_2016} describes 13 ranking-based semantics, it is only these gradual semantics which allow for an initial weight to be assigned to an argument.

\begin{definition}[Gradual Semantics]
\label{def:grad_sem}
A gradual semantics $\sigma$ is a function that associates to each weighted argumentation graph $\AF=\langle \A,\D,w \rangle$, a scoring function $\sigma^\AF : \A \to [0,1]$ that provides an acceptability degree to each argument.
In this paper, we consider the gradual semantics $\sigma_x$, for $x \in \{TB,IS, MB, CB, HC\}$, defined as follows.
\begin{itemize}
    \item The \emph{trust-based} semantics $\sigma_{TB}$ \cite{TB-sem} is defined such that the acceptability degree of an argument $a\in \A$ is  $\sigma_{TB}^\AF(a)  = TB_\infty(a)$, where
    $TB_i(a) = \frac{1}{2} \cdot TB_{i-1}(a)+\frac{1}{2} \cdot  \min(w(a),1-\max\limits_{b \in \Att(a)} TB_{i-1}(b))$ and for all $b \in \A$, $TB_0(b) = w(b)$.
    
    \item The \emph{iterative-schema} semantics $\sigma_{IS} $ \cite{IS-sem} is defined such that the acceptability degree of an argument $a\in \A$ is $\sigma_{IS}^\AF(a) = IS_\infty(a)$, where $IS_i(a)=(1-IS_{i-1}(a)) \cdot \min(\frac{1}{2},1-\max\limits_{b \in \Att(a)} IS_{i-1}(b))+IS_{i-1}(a) \cdot \max(\frac{1}{2},1-\max\limits_{b \in \Att(a)} IS_{i-1}(b))$ and for all $b \in \A, IS_0(b) = w(b)$.
    
    \item The \emph{weighted max-based} semantics $\sigma_{MB}$ \cite{AMGOUD2022103607} is defined such that the acceptability degree of an argument $a\in \A$ is $\sigma_{MB}^\AF(a) = MB_\infty(a)$, where $MB_i(a)=\frac{w(a)}{1+\max\limits_{b \in \Att(a)} MB_{i-1}(b)}$ and for all $b \in \A, MB_0(b) = w(b)$
    %if $\Att(a) \neq \emptyset$, and $w(a)$ otherwise
    .

    \item The \emph{weighted card-based} semantics $\sigma_{CB}$ \cite{AMGOUD2022103607} is defined such that the acceptability degree of an argument $a\in \A$ is $\sigma^\AF_{CB}(a) = CB_\infty(a)$ where
    $CB_i(a)=\frac{w(a)}{1+|\Att^*(a)|+ \frac{\sum\limits_{b \in \Att^*(a)} CB_{i-1}(b)}{|\Att^*(a)|}}$, for all $b \in \A, CB_0(b) = w(b)$, and $\Att^*(a) = \{ b \in \Att(a) \mid w(b)>0\}$ if $\Att^*(a) \neq \emptyset$ and $w(a)$ otherwise.
    
    \item The \emph{weighted h-categorizer} semantics $\sigma_{HC} $\cite{AMGOUD2022103607} is defined such that the acceptability degree of an argument $a \in \A$ is $\sigma_{HC}^\AF(a) = HC_\infty(a)$ where
    $HC_i(a)=\frac{w(a)}{1+\sum\limits_{b \in \Att(a)} HC_{i-1}(b)}$ and for all $b \in \A, HC_0(b) = w(b)$
    %if $\Att(a) \neq \emptyset$ and $w(a)$ otherwise
    .
\end{itemize}
\end{definition}

With the exception of $\sigma_{IS}$, the ranking on arguments is obtained from the acceptability degree assigned to them. For $\sigma_{IS}$, the semantics returns those arguments whose acceptability degree is set to 1.
As usual, for every $a,b \in \A$, we write $a \succ b$ iff $a \succeq b$ and $b \not \succeq a$, $a \preceq b$ iff $a \not \succ b$, $a \prec b$ iff $a \not \succeq b$, and $a \simeq b$ iff $a\preceq b$ and $a \succeq b$. 

\begin{example} \label{ex:semantics}
Let $\AF = \langle \A, \D, w \rangle$ be a WAF, where $\A = \{ a_0, a_1, a_2, a_3 \}, \D = \{ (a_0, a_2),(a_1, a_1),(a_1, a_2),(a_2, a_2),$ $(a_3, a_2) \}$, $w(a_0) = 0.43, w(a_1) = 0.39, w(a_2) = 0.92$, and $w(a_3) = 0.3$. The WAF is represented in Figure \ref{fig:ex1} and the acceptability degrees for the gradual semantics of Definition \ref{def:grad_sem} are shown in Table \ref{tab:ex1}. 

\begin{table}[!h]
    \centering
    \renewcommand{\arraystretch}{1.3}
    \begin{tabular}{|c|c|c|c|c|c|}
    \hline
         & $a_0$& $a_1$& $a_2$& $a_3$& Argument ranking \\
         \hline
         $\sigma^\AF_{TB}$& 0.43 & 0.39 & 0.50 & 0.30 & $a_3 \prec a_1 \prec a_0 \prec a_2$\\
         \hline
         $\sigma^\AF_{IS}$& 1.00 & 0.50 & 0.00 & 1.00 & $a_2 \prec a_1 \prec a_0 \simeq a_3$\\
         \hline
         $\sigma^\AF_{MB}$& 0.43 & 0.30 & 0.58 & 0.30 & $a_1 \simeq a_3 \prec a_0 \prec a_2$\\
         \hline
         $\sigma^\AF_{HC}$& 0.43 & 0.30 & 0.38 & 0.30 & $ a_1 \simeq a_3 \prec a_2 \prec a_0$\\
         \hline
         $\sigma^\AF_{CB}$& 0.43 & 0.18 & 0.17 & 0.30 & $ a_2 \prec a_1 \prec a_3 \prec a_0$\\
         \hline
    \end{tabular}
    \caption{Acceptability degrees of the arguments from Figure \ref{fig:ex1}}
    \label{tab:ex1}
\end{table}

\begin{figure}
\centering
\begin{tikzpicture}[scale=0.1]
\tikzstyle{every node}+=[inner sep=0pt]
\draw [black] (18.5,-39.5) circle (3);
\draw (18.5,-39.5) node {$a_0$};
\draw [black] (32.2,-28.1) circle (3);
\draw (32.2,-28.1) node {$a_1$};
\draw [black] (32.2,-39.5) circle (3);
\draw (32.2,-39.5) node {$a_2$};
\draw [black] (45.8,-39.5) circle (3);
\draw (45.8,-39.5) node {$a_3$};

\draw [black] (21.5,-39.5) -- (29.2,-39.5);
\fill [black] (29.2,-39.5) -- (28.4,-39) -- (28.4,-40);
\draw [black] (30.877,-25.42) arc (234:-54:2.25);
\fill [black] (33.52,-25.42) -- (34.4,-25.07) -- (33.59,-24.48);
\draw [black] (32.2,-31.1) -- (32.2,-36.5);
\fill [black] (32.2,-36.5) -- (32.7,-35.7) -- (31.7,-35.7);
\draw [black] (33.523,-42.18) arc (54:-234:2.25);
\fill [black] (30.88,-42.18) -- (30,-42.53) -- (30.81,-43.12);
\draw [black] (42.8,-39.5) -- (35.2,-39.5);
\fill [black] (35.2,-39.5) -- (36,-40) -- (36,-39);
\end{tikzpicture}
\caption{Graphical representation of a WAF}
\label{fig:ex1}
\end{figure}
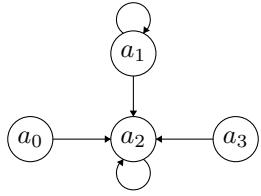

\end{example}

We note that the semantics described above are able to operate on cyclic graphs. Semantics such as DF-Quad \cite{rago_discontinuity-free_2016}, while popular, are designed to operate on acyclic graphs only, and we therefore ignore them in this work.

\subsection{The Bisection Method}
The algorithm we describe in Section \ref{sec:inverseProb} requires us to find the roots of a continuous function. While many techniques for doing so exist \cite{dekker1969finding,brent2013algorithms,newton-raphson}, the bisection method is easily understood and numerically stable, and is therefore used in our experiments. 
Note that more advanced root finding methods could be used within our approach with no loss of generality.

\begin{algorithm}
\begin{algorithmic}
\Function{Bisect}{$f,\alpha,\beta,\varepsilon$}
\State $\mu \gets \frac{\alpha+\beta}{2}$
\State \textbf{if} {$|f(\mu)|< \varepsilon$} \textbf{then } \Return $\mu$
\State \textbf{if} {$f(\mu)>0$} \textbf{then }\Return \Call{Bisect}{$f,\mu,\beta,\varepsilon$}
\State \textbf{else} \Return \Call{Bisect}{$f,\alpha,\mu,\varepsilon$}
\EndFunction
\end{algorithmic}

\caption{The bisection method.}
\label{alg:bisection}
\end{algorithm}

Algorithm \ref{alg:bisection} details the bisection method. As input, the method takes in a function $f$, a tolerance $\epsilon$, and upper and lower bound values ($\alpha$ and $\beta$ respectively), such that $f(\beta)<0<f(\alpha)$. A single iteration of the algorithm identifies the midpoint $\mu=(\alpha+\beta)/2$. If $f(\mu)>0$, $\alpha$ is set to $\mu$; if $f(\mu)<0$, $\beta$ is set to $\mu$, tightening the upper and lower bounds. The process then repeats until the absolute value of the image of the midpoint is sufficiently small, i.e., $|f((\alpha+\beta)/2)|<\epsilon$. 
Note that one can choose to also stop when the distance between $\alpha$ and $\beta$ is small.
The number of iterations required to achieve an error $\epsilon$ is bounded by $\lceil \log_2((\mid\alpha-\beta\mid)/\epsilon) \rceil$. 
Note that for the bisection method to work correctly and return a unique root, the function $f$ must be continuous and monotonic in the interval $[\alpha,\beta]$.

\section{The Inverse Problem} \label{sec:inverseProb}

Our aim in this work is to identify a set of initial weights to obtain some desired final ranking on arguments. More formally, we take as input: (1) an \textit{unweighted} argumentation framework $\langle \A,\D \rangle$, (2) a gradual semantics $\sigma$, and (3) a desired preference relation $\succeq \subset \A \times \A$. 
Our aim is to find a weighting function $w$ such that in the resultant WAF $\AF=\langle \A,\D,w \rangle$, for all $a,b \in \A, \sigma^\AF(a) \geq \sigma^\AF(b)$ iff $a \succeq b$.

In Sections \ref{sec:phase1} and \ref{sec:phase2}, we describe an algorithm to identify an appropriate weighting function. The proposed algorithm involves two phases. In phase 1, we identify an achievable acceptability degree for an argument, taking into account the desired ranking on arguments. In phase 2, we undertake a search --- using the bisection method --- for the initial weights necessary to achieve this desired acceptability degree.

Before examining the algorithm, we consider several special cases of the inverse problem.
%is either easy, or impossible, to solve.

\subsection{Special Cases}

We begin by considering the trust-based semantics. We note that, if $w(a)<0.5$ for all $a \in \A$, then $\sigma_{TB}^\AF(a)=w(a)$, making the inverse problem trivial to solve in this case.
We note that while such a solution satisfies the inverse problem, it is at odds with the intuition behind trust based semantics as described in \cite{TB-sem}. % I cited the IJCAI 2011 paper, I guess that's what you wanted ;-) 
%As discussed further in Section \ref{}, 
In cases where, for all $a\in \A$, $w(a) \geq 0.5$, the presence of cycles can mean that no solution exists for the inverse problem under the $\sigma_{TB}$ semantics. As an example of this, consider the standard 3-cycle WAF: $\langle \{a,b,c\},\{(a,b),(b,c),(c,a)\},w \rangle$. If $w(a),w(b),w(c) \geq 0.5$, the acceptability degrees of all arguments will be 0.5.

Turning to the $\sigma_{IS}$ semantics, we observe that it was designed to have acceptability degrees converge to either 1, 0.5, or 0. This means that the inverse problem is not always applicable to this semantics as it can only accommodate three levels of acceptability. Moreover, there are rankings which cannot be achieved, e.g. consider the simple WAF: $\langle \{a,b\}, \{ (a,b)\}, w \rangle$, it is not possible to get $a \prec b$ as $\sigma_{IS}^\AF(a) = 1$ and $\sigma_{IS}^\AF(b) = 0$, for any weighting $w$.
%
%Another problem is that our proposed algorithm relies on the fact that the acceptability degree of an argument is always equal or less than its initial weight which is not the case with this semantics (see Example \ref{ex:semantics})

Finally, consider a fully connected graph. We can easily prove the following proposition, which makes the solving the inverse problem on such graphs trivial.

\begin{proposition}
\label{prop:fully-connected}
For a fully connected WAF $\AF=\langle \A, \D, w \rangle$, semantics $\sigma \in \{\sigma_{MB}, \sigma_{CB}, \sigma_{HC}\}$ and any arguments $a,b \in \A$, $\sigma^\AF(a) \geq \sigma^{\AF}(b)$ iff $w(a) \geq w(b)$.
\end{proposition}

Given these special cases, in the remainder of the paper we consider only $\sigma_{MB}, \sigma_{CB}$ and $\sigma_{HC}$. We can trivially solve the inverse problem for fully connected graphs as all the semantics will converge quickly, even in the presence of a significant number of arguments and edges.

\subsection{Phase 1: Computing Acceptability Degrees}
\label{sec:phase1}
We partition the set of arguments $\A$ into a sequence of non-empty sets of arguments $[Ar_0, \ldots, Ar_n]$ such that for any $a,b \in Ar_i$, $0 \leq i \leq n$, $a \simeq b$, and for any $a \in Ar_i, b \in Ar_j$ where $0 \leq i < j \leq n$, $a \succ b$.
Now consider an argument $a \in Ar_0$. For each semantics, we can reason as follows.
\begin{itemize}
    \item $\sigma_{MB}$: Assume that $a$  is attacked by an argument with acceptability degree 1. If $w(a) =1$, its acceptability degree can be at most 0.5.
    \item $\sigma_{CB}$: Assume that $a$ is attacked by $n$ other arguments with degree 1. Then its acceptability degree can be at most $1/(2+n)$. If $a$ is the most attacked argument in $Ar_0$, then all other arguments in $Ar_0$ will have an acceptability degree equal to or greater than this value.
    
    \item $\sigma_{HC}$: Assume that $a$ is attacked by $n$ other arguments with degree 1. Then its acceptability degree can be at most $1/(1+n)$. If $a$ is the most attacked argument in $Ar_0$, then all other arguments will have an acceptability degree equal or greater to this value.
\end{itemize}

We refer to the aforementioned values as the \emph{minimal upper bounds} for the arguments in $Ar_0$, as this is the lowest value we are guaranteed to be able to achieve (with the corresponding semantics) if the arguments in $Ar_0$ have an initial weight of 1.
Similarly, the \emph{maximal upper bound} for arguments in $Ar_0$ is 1, achievable if all attackers of arguments in $Ar_0$ have an acceptability degree of 0.
The idea is to make sure that all the arguments from $Ar_0$ have acceptability degree in the interval $[mup, 1]$, where $mup$ is the minimal upper bound corresponding to the semantics in question, e.g.\ all the arguments from $Ar_0$ are within $[\frac{1}{1+n}, 1]$ for $\sigma_{HC}$.

Now, consider $Ar_1$. If the maximal upper bound of the acceptability degree of these arguments is lower than the minimal upper bound for the arguments in $Ar_0$, then we will comply with our desired ranking on arguments. To achieve this, we set the initial weights of arguments in $Ar_1$ to (just below) the minimal upper bound of $Ar_0$. We can repeat this, computing initial weights, and concomitant maximal upper bounds for $Ar_i$ by considering the minimal upper bounds of $Ar_{i-1}$. Algorithm \ref{alg:mub} describes this process. Note that a small constant $\zeta$ is added to the denominator in all cases to ensure that the minimal upper bound is still reduced for the special case where all arguments in some $Ar_i$ are unattacked.

\begin{algorithm}
\begin{algorithmic}
\Function{ComputeBounds}{$[],\_,\_$}
\State \Return $\{\}$
\EndFunction\medskip
\Function{ComputeBounds}{$[Ar_0, \ldots, Ar_n],max,\sigma$}
\Switch{$\sigma$}
\Case{$\sigma_{MB}$} 
  $min \gets max/(1+max+\zeta)$
\EndCase
\Case{$\sigma_{HC}$}
  $min \gets max/(1+\max\limits_{a \in Ar_0} |\Att(a)|+\zeta)$
\EndCase  
\Case{$\sigma_{CB}$}
  $min \gets max/(2+\max\limits_{a \in Ar_0} |\Att(a)|+\zeta)$
\EndCase
\EndSwitch
\State \Return $\{(Ar_0,min)\} \cup$ \Call{ComputeBounds}{$[Ar_1, \ldots, Ar_n],min, \sigma$}
\EndFunction
\medskip

\Function{ComputeBounds}{$[Ar_0, \ldots, Ar_n], \sigma$}

\Return \Call{ComputeBounds}{$[Ar_0, \ldots, Ar_n],1,\sigma$}
\EndFunction
\end{algorithmic}
\caption{Computing arguments' minimal upper bounds} \label{alg:mub}
\end{algorithm}

\subsection{Phase 2: Finding the Initial Weights}
\label{sec:phase2}
Having identified appropriate minimum upper bounds for all $Ar_0$ to $Ar_n$, we now turn our attention to finding initial weights for each argument in these sets so as to have that argument's acceptability degree equal to the corresponding set's minimum upper bound. By doing this, we obtain our desired ranking on arguments.

Our approach to achieving this involves picking an argument and modifying its initial weight (using the bisection method), causing it to approach its minimum upper bound value. We then pick another argument and repeat this process, until all arguments reach their desired values. There are several choices we must consider, and optimisations possible, when instantiating this approach. The most obvious choices we face revolve around selecting an argument for modification, and the decision of how much to modify the selected argument by.
Myriad strategies for argument selection are possible, and in this work we consider 5 simple strategies:

\begin{itemize}
    \item[$S1:$] Select more preferred arguments for modification first. The rationale here is that such arguments have higher acceptability degrees, and fixing their values will cause fewer perturbations in the remainder of the process.
    \item[$S2:$] Select less preferred arguments for modification first. Such arguments, with their small degree, would have little influence on the network.
    \item[$S3:$] Select arguments further from their target degree first. By selecting arguments with the largest error first, we may perturb the network less.
    \item[$S4:$] Select arguments nearest their target degree first. These, due to needing only minor perturbations, would have minimal effect on the rest of the argumentation system.
    \item[$S5:$] Pick arguments at random. This is the baseline strategy.
\end{itemize}

Observe that additional strategies can be used, e.g. picking arguments with most, or fewest attackers, or which attack most or fewest arguments, first. We leave consideration of such strategies to future work.

It is important to note that the bisection method is only guaranteed to work for a function with a single variable, and the selection of an appropriate strategy is therefore critical to our algorithm's success. As discussed in Section \ref{sec:eval}, some of these strategies work much better than others, but we are unable to provide an analytical proof of correctness for any of the strategies.
%\footnote{We note some cases where certain strategies perform poorly within our supplementary material.} 

With regards to how much we should modify a selected argument, we could do so until it is within some tolerance $\epsilon$ of its acceptability degree, or until a certain number of iterations of the bisection method have been carried out. The rationale behind the second approach is that it allows us to respond to changes in acceptability degrees of other arguments due to our modifications more rapidly than if we modify only a single argument at a time.

If $d$ is the desired acceptability degree for an argument $a$, then we can use the bisection method to find a new initial weight $w_a$ for $a$ such that $|\sigma^\AF(a)-d| \leq \epsilon$ where $\sigma^\AF(a)$ is the acceptability degree of $a$ in the WAF where the weight of $a$ is now $w_a$.
To apply the bisection method we need to also identify an initial upper and lower bound. While we can use the values 1 and 0 for this, we can also identify tighter bounds, leading to a small improvement in performance.
First, consider the lower bound $\alpha$ passed to the bisection method. Since our denominator is at least 1, we can set $\alpha$ to the minimal upper bound. For $\beta$, assume we wish to achieve a minimal upper bound of $m(a)$ for argument $a$, which has $n$ attackers. 
Now consider $\sigma_{MB}$, and assume that the strongest attacker has acceptability degree 1. We have that $m(a)=w(a)/(2+\zeta)$ and so can set $\beta$ to $\min\{(2+\zeta) \cdot m(a),1\}$. Using this idea, for $\sigma_HC$, we can set $\beta$ to $\min\{m(a) \cdot (1+n+\zeta),1\}$, and for $\sigma_{CB}$ to $\min\{(2+n+\zeta) \cdot m(a),1\}$.

From a practical point of view, observe that the target acceptability degree computed in Phase 1 may be very small. The stopping condition of our bisection method should therefore use a relative error $|(\alpha + \beta)/2-m(a)|/m(a)<\epsilon$ rather than an absolute error. Since we evaluate acceptability degrees as part of the bisection method, we can also terminate our algorithm early if the acceptability degrees returned in the evaluation match our desired ranking, even if they have not yet converged to the desired minimum upper bound.

\section{Algorithm Properties}\label{sec:prop}
We now examine several properties of our approach and the underlying semantics, identifying necessary conditions over the latter which are needed for the former to work. Given the iterative nature of the underlying semantics, proving that some of these properties hold is difficult, and in Section \ref{sec:eval}, we carry out an empirical evaluation which strongly suggests that the $\sigma_{MB},\sigma_{HC}$ and $\sigma_{CB}$ semantics respect these properties. Properties which we are able to demonstrate are identified as propositions, with associated proofs in the supplementary material, while those we are unable to analytically demonstrate are labelled as conjectures.

The first property we consider involves the weights obtained in Phase 1 (Section \ref{sec:phase1}). We need to demonstrate that the computed weights are achievable. While we can easily demonstrate that the computed weight is achievable in isolation, doing so for the entire system is more difficult.

\begin{conjecture}[Weighting Validity]
\label{weight-existence}
For any unweighted argumentation graph $\langle \A, \D \rangle$ and $\sigma \in \{ \sigma_{MB}, \sigma_{HC} , \sigma_{CB} \}$, there is a
%\todo{unique???} 
weighting function $w$ such that for all $0 \leq i \leq n$, for all $a \in Ar_i$, $\sigma^\AF(a)$ is equal to its minimum upper bound (as computed by Algorithm \ref{alg:mub}), where $\AF = \langle \A , \D, w \rangle$.
\end{conjecture}

In Phase 2, for the bisection method to operate, we must demonstrate that our semantics is continuous  (otherwise we may be unable to converge to a solution); and that these changes are (strongly) monotonic (as otherwise we may have any number of solutions). This also implies that our solution satisfies uniqueness (though uniqueness does not imply monotonicity).

\begin{property}[Uniqueness]
\label{prop:uniqueness}
Given two WAFs $\AF=\langle \A, \D, w \rangle, \AF'=\langle \A, \D,w'\rangle$ and some $a \in \A$ such that $w(a)\neq w'(a)$ and for all $ b \neq a \in \A, w(b)=w'(b)$. It holds that $\sigma^\AF(a) \neq \sigma^{\AF'}(a)$, for $\sigma \in \{\sigma_{MB},\sigma_{CB},\sigma_{HC}\}$.
\end{property}

\begin{property}[Continuity]
%A function $f(x)$ satisfies continuity iff $\lim_{x \to c} f(x)=f(c)$.
%A gradual semantics $\sigma$ satisfies continuity iff for any two WAFs $\AF = \langle \A, \D,w \rangle$, $\AF' = \langle \A, \D,w' \rangle$, and $a\in \A$ s.t.\ $w'(a)=w(a)+\delta$ and for any $b \in \A \setminus \{a\}$, $w(b)=w'(b)$, then $\lim\limits_{\delta \to 0} \sigma^{\AF'}(a)=\sigma^{\AF}(a)$.

A gradual semantics $\sigma$ satisfies continuity iff for any WAF $\AF = \langle \{ a_1, a_2, \dots, a_n\},\D,w\rangle$, $X^\AF = (\sigma^\AF(a_1), \sigma^\AF(a_2), \dots, \sigma^\AF(a_n))$, we can find $\AF' = \langle \{ a_1, a_2, \dots, a_n\},\D,w'\rangle$ (with unbounded weights) such that there is at least one $a \in \A$ s.t.~$w(a) \neq w'(a)$ and  $|X^{\AF'}-X^{\AF}|<\delta$ for any positive $\delta$.
We say that $\sigma$ satisfies bounded continuity iff it satisfies continuity and the initial weights for $\AF'$ are restricted to $[0,1]$.
\end{property}

\begin{property}[Strong Monotonicity]
A gradual semantics $\sigma$ satisfies strong monotonicity iff for any two WAFs $\AF=\langle \A, \D, w \rangle, \AF'=\langle \A, \D,w'\rangle$ for which there is some $a \in \A$ such that $w(a) = w'(a) + \delta, \delta > 0$ and for all $ b \in \A \setminus \{a\}, w(b)=w'(b)$, it holds that $\sigma^\AF(a) > \sigma^{\AF'}(a)$.

%A function $f$ exhibits \emph{strong monotonicity} iff for any $a>b$, $f(a)>f(b)$. \todo{So we need to show $w(a)f(a)<(w(a)+\delta)f(a+\delta)$ for positive $\delta$. Which could occur if delta grows faster than f(a) shrinks.}
\end{property}

\noindent This in turn yields the following proposition.

\begin{proposition}
\label{prop:necessary}
A gradual semantics $\sigma$ which does not satisfy strong monotonicity or bounded  continuity could have multiple, or no solutions to the inverse problem. In other words, both are necessary conditions for a unique solution for the inverse problem to exist.
\end{proposition}

%Without monotonicity we can have multiple roots and our algorithm may not converge. 
%Without continuity, we may require a value which we cannot achieve.
%TODO: provide examples which demonstrate this.

We conjecture that $\sigma_{MB}, \sigma_{HC}$ and $\sigma_{CB}$ meet  these conditions. We are unable to demonstrate strong monotonicity and bounded continuity though we show uniqueness and continuity for them in our supplementary material). The empirical evaluation suggests our approach operates successfully.

\begin{proposition}
\label{conjecture:necessary}
The gradual semantics $\sigma$ satisfies continuity and uniqueness, for $\sigma \in \{ \sigma_{MB}, \sigma_{HC}, \sigma_{CB}\}$ .
\end{proposition}

\begin{conjecture}
\label{conjecture:strongmono}
The gradual semantics $\sigma$ satisfies strong monotonicity and bounded continuity for $\sigma \in \{ \sigma_{MB}, \sigma_{HC}, \sigma_{CB}\}$ .
\end{conjecture}

%\begin{property}[Weight Existence]
%A gradual semantics $\sigma$ satisfies weight existence iff for any unweighted argumentation graph $\langle \A, \D \rangle$ and any ranking $\succeq$ on $\A$, there exists a weighting function $w$ such that for all $a,b \in \A, (a,b) \in \succeq$ iff $\sigma^\AF(a) \geq \sigma^\AF(b)$, where $\AF = \langle \A, \D,w \rangle$.
%\end{property}

\section{Evaluation} \label{sec:eval}

We evaluated each of the strategies discussed in Section \ref{sec:phase2} over directed scale-free, small world (Erdos-Renyi), and complete graphs of different sizes (number of arguments)\footnote{Source code for our algorithm and evaluation can be found on GitHub at \url{https://github.com/jhudsy/numerical_inverse}.}. As part of our evaluation, we ran 10, 100 and 2000 iterations of the bisection method for each argument before using a strategy to pick the next desired argument. Table \ref{tab-parameters} describes the remaining parameters used in our evaluation. 

We created a simple target preference ordering for our experiments, randomly placing each argument within the graph into one of 5 levels of preference. This meant that in all cases, at least some arguments had equal desired preference levels.

Our experiments evaluated the runtime of the different strategies, the number of times the bisection method was invoked, and the number of times the total iterations required exceeded the permitted maximum number of iterations. Given the number of dimensions across which our evaluation took place, we present only a subset of our results here; full results can be found in the supplementary material.

\begin{table}
\centering
\begin{tabular}{|l|l|}
  \hline
  $\zeta$ & 1 \\
  Graph Size   & 10, 20, \ldots, 150 \\
  Runs per graph size & 15 \\
  Erdos-Renyi probability & 0.1,0.3,0.5,0.7 \\
  Maximum relative error   & 0.001 \\
  Bisection method iterations & 10,100,2000 \\
  Bisection method $\epsilon$ & 0.001 \\
  Maximum bisection method calls & 1000 \\%means that for 200 args, each arg visited 5 times?!? For 10 args, 100 visits...
  \hline
\end{tabular}
\caption{Parameters used in our evaluation}
\label{tab-parameters}
\end{table}

\begin{figure*}
    \centering
    \includegraphics[width=0.4\textwidth]{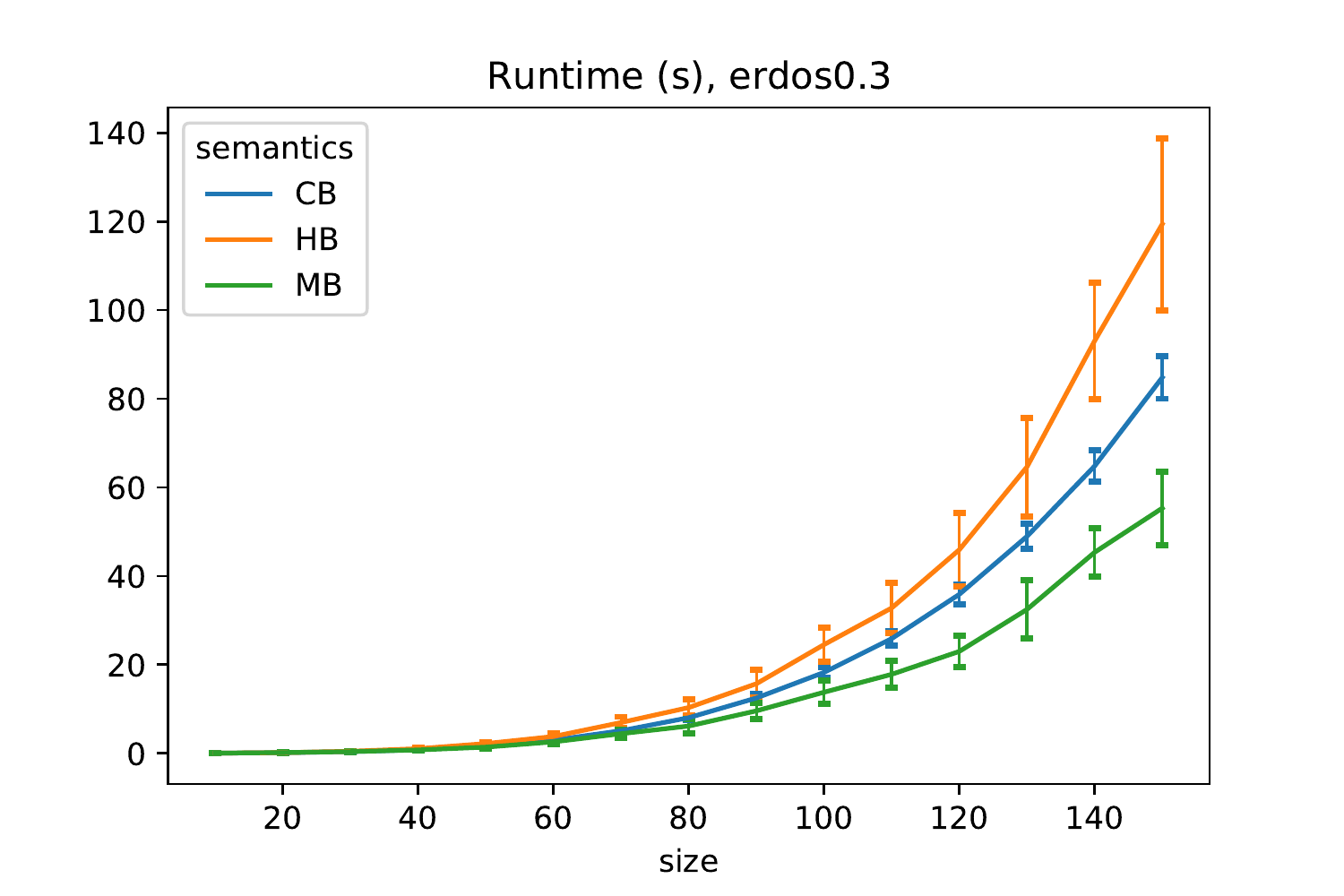}
    \includegraphics[width=0.4\textwidth]{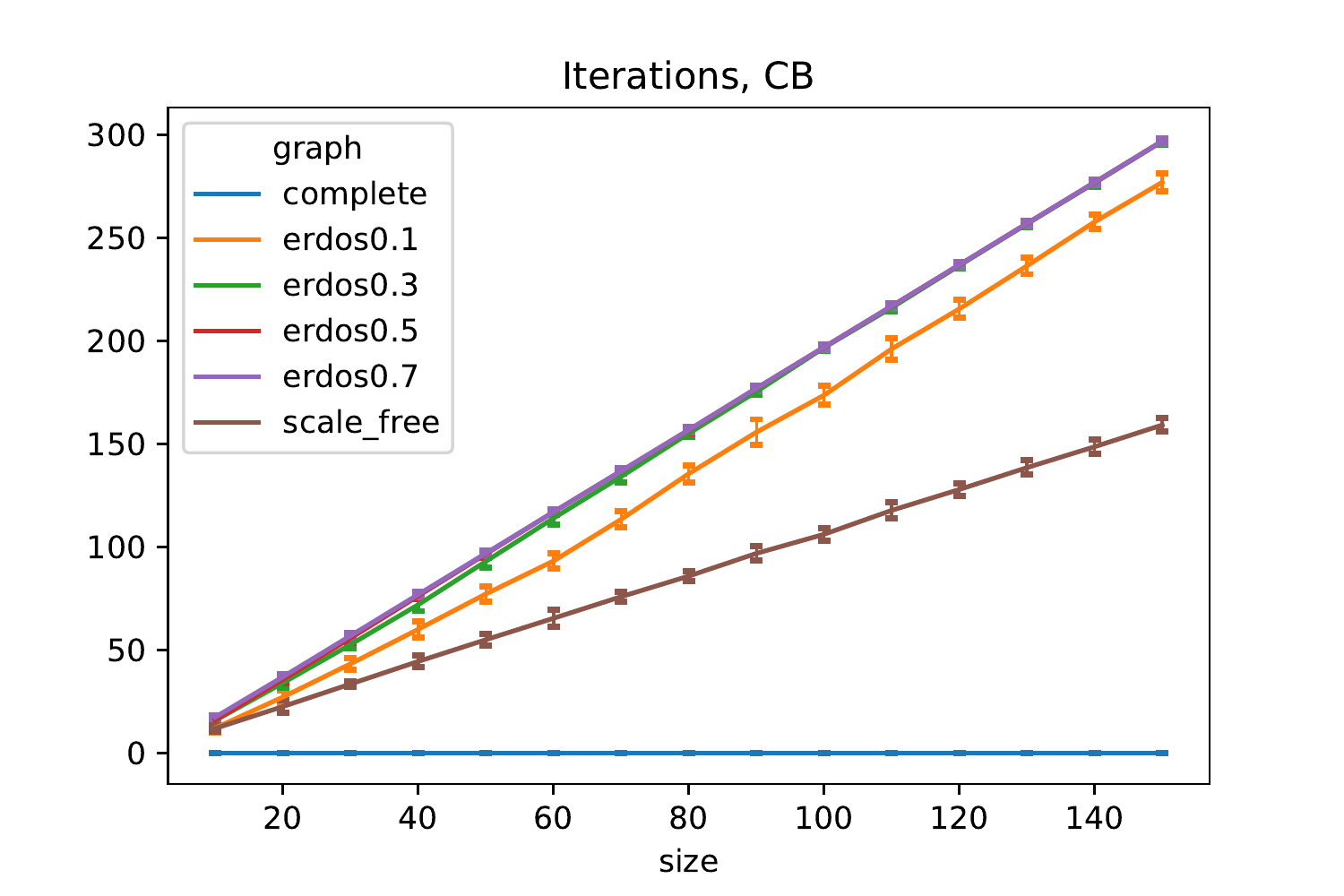}
    \includegraphics[width=0.4\textwidth]{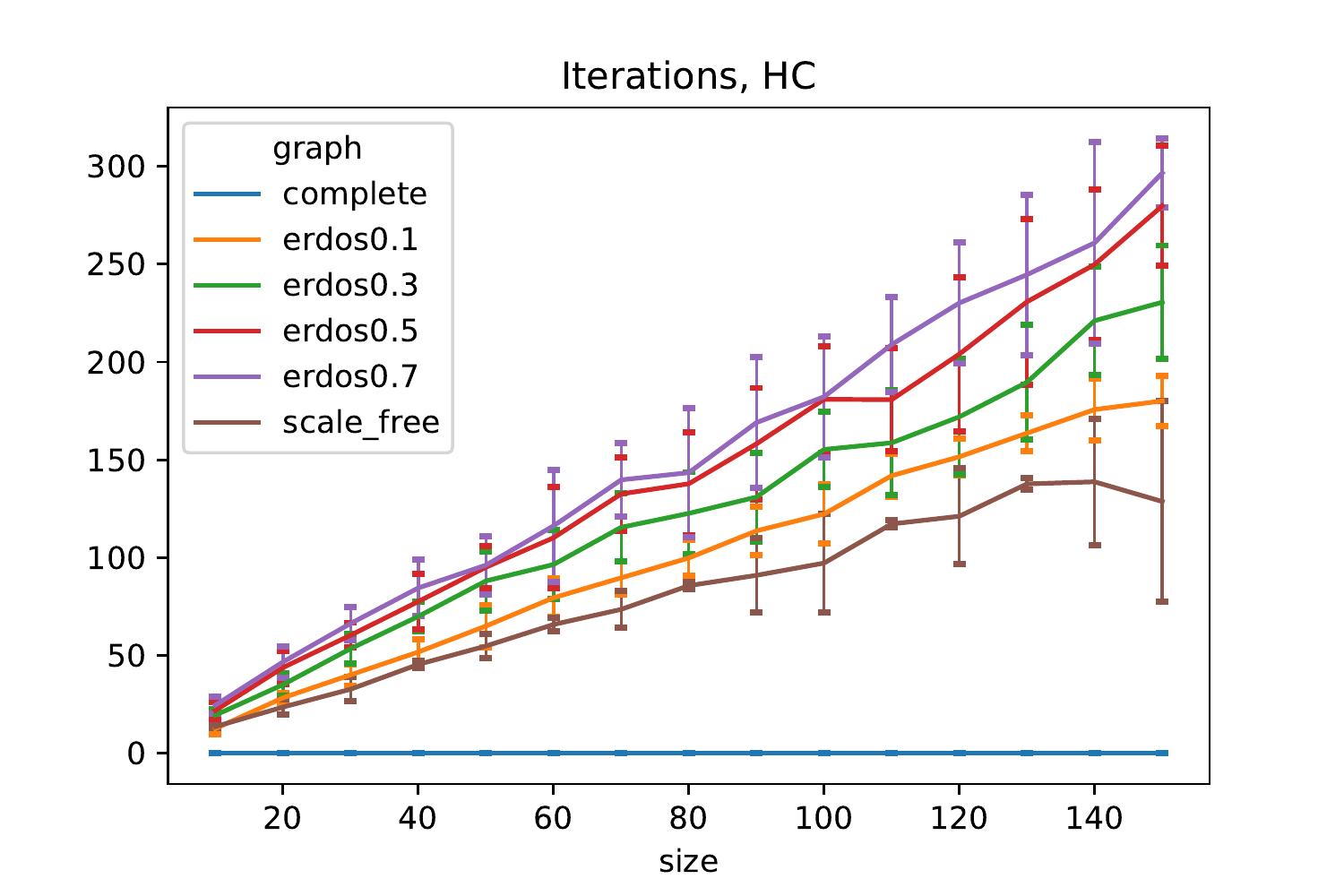} 
    \includegraphics[width=0.4\textwidth]{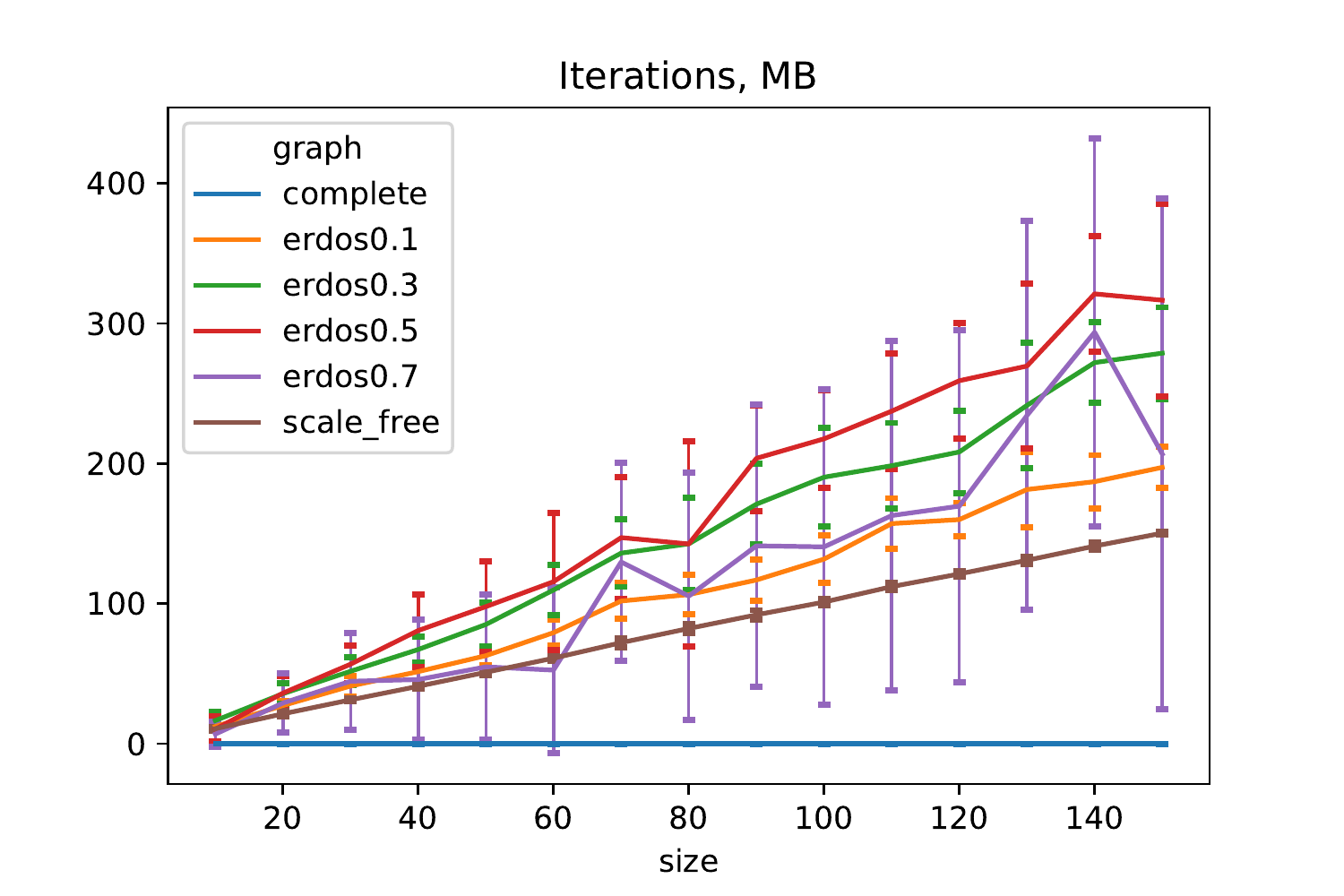}
    \caption{Runtime (in seconds) and number of iterations for the different semantics and graph types}
    \label{fig:eval}
\end{figure*}

Our main criterion for evaluation revolves around the number of times the bisection method was called by our approach. As shown in Figure \ref{fig:eval}, which is representative of the results for most graph topologies, our runtime grows in a super-linear manner, due --- as shown in \cite{AMGOUD2022103607} --- to the increased time taken to evaluate a semantics on larger graphs. The number of bisection method iterations is shown in Figure \ref{fig:eval} for our different semantics and graph types when selecting the next argument based on largest relative error. We observe that this value grows linearly (with a gradient between 1 and 2 depending on topology and semantics) for all semantics considered ($R^2>0.99$ for all cases). This means that arguments are typically only recomputed at most twice before our approach converges. These results not only demonstrate the feasibility of our approach for large argumentation graphs, but also highlight the effectiveness of this specific argument selection strategy. We also note that there is little variance in our results for $CB$. We believe that this is due to the extra $|Att|$ term in the semantics; this term overwhelms the term which depends on other arguments' final degrees, making the result more dependent on the topology of the graph than in the case of the other semantics.

Due to space, we have omitted detailed results about several other aspects of our approach. In summary,

\begin{itemize}
    \item From all argument selection strategies, only selecting arguments from largest to smallest relative error resulted in always finding a solution to the inverse problem.
    \item Allowing for partial convergence (via fewer iterations per argument before moving to the next one) decreased performance, often failing to find a solution.
    \item Optimising the $\alpha$ and $\beta$ bounds (c.f. Section \ref{sec:phase2}) had almost no influence on runtime. This is not  surprising due to the speed at which the bisection method converges.
\end{itemize}

Note that in the absence of equivalent arguments with the same acceptability degree, one could allow for early termination without getting to the target acceptability degrees, but that was not investigated in the current paper.

\section{Conclusion \& Discussion} \label{sec:discussion}

%CONCLUSION
In this paper we considered the inverse problem for a weighted argumentation framework. We demonstrated that a solution to this inverse problem exists for at least one family of semantics, and that a solution does not exist for at least some semantics. We then described an algorithm to solve the inverse problem, and empirically evaluated its performance. Our results show the viability of our approach. When selecting arguments for initial weight perturbation based on relative error, each argument is typically perturbed at most twice (depending on semantics and graph topology).

%Discussion
Our approach was able to find weightings over all evaluated graphs and semantics, suggesting that $HC, CB$ and $MB$ meet all the requirements described in Section \ref{sec:prop}. While we have been unable to analytically prove many of our results, relying instead on an empirical analysis, searching for such analytic proofs forms a critical avenue of future work.

No work has --- to our knowledge --- explicitly considered the inverse problem applied to gradual semantics as described in this paper, but several works have examined related concepts under different guises. Work on the epistemic approach to probabilistic argumentation \cite{hunter17probabilistic} describes families properties which probabilistic argumentation semantics should satisfy. Selecting a set of these properties constrains the possible probabilities which arguments can have. A similar strand of work in the context of fuzzy argumentation \cite{wu16fuzzy} allows one to calculate legal ranges of fuzzy degrees for arguments based on initial weights assigned to arguments and the semantics underpinning the fuzzy argumentation system.
Another strand of somewhat related work comes from the area of argumentation dynamics. Work here examines what arguments or attacks should be introduced to strengthen or weaken an argument, in a manner somewhat analogous to our changing of an initial argument weight.

There has been some work on sensitivity analysis within argumentation \cite{tang2016markovargumentationrandom}. This work considers whether (small) changes in argument weights will affect the conclusions that can be drawn from an argumentation framework. 

We are extending this research in several directions as part of our current and future work. The results reported on in this paper are a first step towards our long-term goal to provide a formal analysis of sensitivity to initial weights in $MB, CB$ and $HC$ style semantics.

The conditions specified in Prop.~\ref{prop:necessary} are necessary for our algorithm to operate. As mentioned above, we are still investigating whether these conditions are also sufficient, or whether additional properties need to be identified. Once this is done, we will be able to categorise other weighted semantics unrelated to those discussed in the current work (e.g., the constellation-based probabilistic semantics \cite{li11probabilistic}) and consider whether our approach can be applied to them. 

The efficiency of our approach suggests that the underlying problem can be solved analytically. A final strand of future work involves searching for such an analytical solution. We note that the different equations describing the respective semantics would require fine-tuning such a solution to the individual semantics, while the bisection method proposed in the current work can be applied more generally.

%\section{Conclusions}

\newpage
\bibliographystyle{abbrv}
\bibliography{biblio}

\appendix 

\newpage 

\section{Appendix - Proofs}

\paragraph{Proposition \ref{prop:fully-connected}} For a fully connected WAF $\AF=\langle \A, \D, w \rangle$, semantics $\sigma \in \{\sigma_{MB}, \sigma_{CB}, \sigma_{HC}\}$ and any arguments $a,b \in \A$, $\sigma^\AF(a) \geq \sigma^{\AF}(b)$ iff $w(a) \geq w(b)$.

\medskip

\begin{proof}
Let $\sigma \in \{ \sigma_{MB}, \sigma_{CB}, \sigma_{HC}\}$ and $\AF = \langle \A, \D, w \rangle$ be a fully connected WAF (including self attacks for every arguments).
As a result, for every $a \in \A$, $\sigma^\AF(a) = \frac{w(a)}{1+ X_\AF^\sigma}$, where $X_\AF^\sigma$ is a positive constant which depends on the structure of the WAF and the semantics (see Definition \ref{def:grad_sem}).
Now, consider two arguments $a,b$ such that $w(a) \leq w(b)$, it holds that $\sigma^\AF(a) \leq \sigma^\AF(b)$ (and vice-versa).
\end{proof}

\paragraph{Proposition \ref{prop:necessary}}

A gradual semantics $\sigma$ which does not satisfy strong monotonicity or bounded  continuity could have multiple, or no solutions to the inverse problem. In other words, both are necessary conditions for a unique solution for the inverse problem to exist.

\medskip

\begin{proof}
To show that multiple or zero solutions exist, we provide two examples.
\begin{itemize}
    \item Consider the semantics $\sigma_{NM}$ such that for all WAFs $\AF = \langle \A, \D, w \rangle$ and $a \in \A, \sigma_{nm}^\AF(a)= \min (-2w(a)^2+w(a), 0)$. It is clear that $\sigma_{NM}$ does not satisfy strong monotonicity. 
    In the case of a WAF with a single argument $a$. If we wish to obtain an acceptability degree of 0 to $a$, we need to assign an initial weight of $0$ or any number in $[0.5, 1]$ to $a$, demonstrating multiple solutions.

    \item The $\sigma_{IS}$ semantics does not satisfy continuity. For example, we cannot achieve an acceptability degree of $0.9$ for any argument, demonstrating that no solution exists to the inverse problem.
\end{itemize}
\end{proof}

\paragraph{Proposition \ref{conjecture:necessary} (Uniqueness)} Given two WAFs $\AF=\langle \A, \D, w \rangle, \AF'=\langle \A, \D,w'\rangle$ for which there is some $a \in \A$ such that $w(a)\neq w'(a)$ and for all $ b \neq a \in \A, w(b)=w'(b)$. It holds that $\sigma^\AF(a) \neq \sigma^{\AF'}(a)$, for $\sigma \in \{\sigma_{MB},\sigma_{CB},\sigma_{HC}\}$.

\medskip

\begin{proof}
Assume an arbitrary WAF $\AF = \langle \A, \D,w \rangle$ such that $a \in \A$ and $\sigma \in \{\sigma_{MB},\sigma_{CB},\sigma_{HC}\}$.
%We denote by $D(a)$ the final degree of $a$.
Assume there is a path from $a$ to one of its attackers, we have that $\sigma^\AF(a)=w(a) / f_a(\sigma^\AF(a))$, where $f_a$ is a function that depends on the WAF and the semantics considered. 
For instance, if $\AF = \langle \{a_0, a_1\}, \{ (a_0, a_1), (a_1, a_0)\}, w \rangle$ and $\sigma = \sigma_{HC}$, then $f_{a_0}(x) = 1 + (w(a_1)/(1+x))$.
If there is no path from $a$ to one of its attackers, $f_a$ is a constant function.

Now, consider $\AF'=\langle \A, \D,w'\rangle$, where $ w'(a) = w(a) + \delta, \delta \neq 0$ and for all $ b \in \A \setminus \{a\}, w(b)=w(a)$.
%We wish to show that the final degree is different if we have an identical graph with $w(a)$ perturbed by $\delta$.
Let us prove the proposition by contradiction. Assume it is false, i.e., that we have the $\sigma^\AF(a) = \sigma^{\AF'}(a)$. Then we must have $w(a) / f_a(\sigma^\AF(a))=(w(a)+\delta) / f_a(\sigma^{\AF'}(a))$, which can only hold if  $\delta=0$, this is a contradiction.
\end{proof}

\paragraph{Proposition \ref{conjecture:necessary} (Continuity)} The semantics $\sigma \in \{ \sigma_{CB}, \sigma_{HC}, \sigma_{MB}\}$ satisfies continuity.

\medskip

\begin{proof}
In this proof, we first focus on the $\sigma_{HC}$ semantics. 
We will denote by $\Adj$ the square adjacency matrix of a WAF $\AF = \langle \{a_1, a_2, \dots, a_n\}, \D , w \rangle$ and $\HC_t = (HC_t(a_1), HC_t(a_2), \dots, HC_t(a_n) )^{T}$ the column vector of intermediate scores at step $t$ for all arguments.

It holds that for all $a_i \in \A$, ${HC}_{t+1}(a_i) = w(a_i) / ([\Adj \cdot \HC_t]_i + 1)$, where $[\Adj \cdot \HC_t]_i$ is the i-th element of the column vector $\Adj \cdot \HC_t$. At the equilibrium point, it holds that $\sigma_{HC}^\AF(a_i) = {HC}_\infty(a_i) = w(a_i) / ([\Adj \cdot \HC_\infty]_i + 1)$.
The evolution of the semantics is described by: $$\HC_{t+1}=F(\HC_{t})=w(a_i) / ([\Adj \cdot \HC_t]_i + 1),$$ 
\noindent
with $F: [0,1]^n \to [0,1]^n$.

%Let us consider the function $F: [0,1]^n \to [0,1]^n$ such that $F((x_1, x_2, \dots, x_n)) = (HC^1(x_1, x_2, \dots, x_n), \dots, HC^n(x_1, x_2, \dots, x_n))$, where $HC^i(x_1, x_2, \dots, x_n) = HC_\infty(a_i)$ in a WAF where for all $a_j \in \A, HC_\infty(a_j) = x_j$.$
Defining the notation $HC_{t}(a_i) \equiv HC^i_{t}$, the Jacobian matrix of $F$ is:

\[
J = \begin{bmatrix}
\frac{\partial HC^1_{t+1}}{\partial HC^1_{t}} & \frac{\partial HC^1_{t+1}}{\partial HC^2_{t}} & \dots  & \frac{\partial HC^1_{t+1}}{\partial HC^n_t} \\

    \frac{\partial HC^2_{t+1}}{\partial HC^1_t} & \vdots & \vdots &  \frac{\partial HC^2_{t+1}}{\partial HC^n_t} \\
    
    \vdots & \vdots & \vdots  & \vdots \\
    
    \frac{\partial HC^n_{t+1}}{\partial HC^1_t} & \frac{\partial HC^n_{t+1}}{\partial HC^2_t}  & \dots  & \frac{\partial HC^n_{t+1}}{\partial HC^n_t}
\end{bmatrix}
\]

%and the element at row $i$ and column $j$ of the Jacobian at the equilibrium is $J_{ij} = \frac{\partial HC_{\infty}^i}{\partial HC_{\infty}^j}$

%Note also that --- in vector notation --- the evolution of the semantic can be represented as 

%$\HC_{t+1} = \frac{\vec{w}}{1+\mathcal{A}\HC_t}$

We make use of the quotient rule, i.e.\
$\frac{d}{dx} \frac{f(x)}{g(x)}=\frac{f'(x) g(x) -f(x) g'(x)}{g^2(x)}$.
Since $w(a_i)$ is not a function of $HC^j_t$, $\frac{\partial w(a_i)}{\partial HC^j_t}=0$ and $J_{ij}$ reduces to
$- \frac{f(x) g'(x)}{g^2(x)}$, where $f(x) = w(a_i)$ and $g(x) = 1 + [\Adj \HC_t]_i$.
Thus, $J_{ij}=-\frac{ \frac{\partial (1+[\Adj \HC_{t}]_i)}{\partial HC^j_t} \cdot w(a_i)}{(1+[\Adj\HC_t]_i)^2} = -\frac{ \frac{\partial([\Adj\HC_{t}]_i)}{\partial HC^j_t} \cdot w(a_i)}{(1+[\Adj\HC_t]_i)^2} $.

%Note the unitary constant in the derivative, which vanishes at the first derivative, leading to
Now consider the column vector $\Adj\HC_{t}$. $\Adj$ is a matrix containing 1s and 0s, while  $\HC_t$ is a column vector. Then the product $\Adj HC_{t}$ is:

$$\Adj{\HC_{t}}=
\begin{bmatrix}
[\Adj]_{11}[\HC_{t}]_1 + [\Adj]_{12}[\HC_{t}]_2 +\ldots \\
[\Adj]_{21}[\HC_{t}]_1 + [\Adj]_{22}[\HC_{t}]_2 +\ldots \\

\vdots
\\
[\Adj]_{n1}[\HC_{t}]_1 + [\Adj]_{n2}[\HC_{t}]_2 +\ldots

\end{bmatrix}
$$

And so, 

$$\frac{\partial([\Adj \HC_{t}]_i)}{\partial HC^j_t}=[\Adj]_{ij}.$$

Therefore, our Jacobian can be denoted as:

$$J_{ij}=-\frac{[\Adj]_{ij}w(a_i)}{(1+[\Adj\HC_t]_i)^2}$$

The equilibrium point is obtained for $t=\infty$ and it is represented by $\HC_{\infty}$. The Jacobian at the equilibrium point is given by replacing $\HC_t$ in the above by $\HC_{\infty}$.
The calculated Jacobian matrix shows that the evolution dynamics of the semantics is continuous at the equilibrium point, i.e., it has a first derivative at that point. As such, a continuous variation in the initial conditions 
$\HC_t$ in the vicinity of the equilibrium point induces a continuous variation in $\HC_{t+1}$. The equilibrium point can be continuously approached as $t\rightarrow \infty$.
This proof can be adapted in a straightforward manner to deal with the $\sigma_{MB}$ and $\sigma_{CB}$ semantics.

\end{proof}

% \begin{proposition}
% Let $\sigma \in \{ \sigma_{HC}\}$ and $\AF = \langle \A, \D, w \rangle$, $\AF' = \langle \A, \D, w' \rangle$ be two WAFs such that there exists $k \geq 1$ and for all $a \in \A, w(a) = w'(a) / k$. It holds that for every $a,b \in \A, \sigma^\AF(a) \leq \sigma^\AF(b)$ iff $\sigma^{\AF'}(a) \leq \sigma^{\AF'}(b)$
% \end{proposition}

\end{document}